\documentclass[runningheads]{llncs}

\usepackage{cite}
\usepackage{graphicx}
\usepackage{mathtools}
\usepackage{cite}

\usepackage{todonotes}
\usepackage{fancybox}

\usepackage{amsmath,amssymb,amsfonts}
\usepackage{algorithmic}
\usepackage{graphicx}
\usepackage{textcomp}
\usepackage{times}
\usepackage{url}
\usepackage{latexsym}
\usepackage{multirow}
\usepackage{placeins}
\usepackage{pdfpages}
\usepackage{hyperref}
\usepackage{wrapfig}
\usepackage{amssymb}
\usepackage{rotating}
\usepackage{todonotes}
\usepackage{lipsum,booktabs}
\usepackage{multicol}
\usepackage{amsmath}
\usepackage{cleveref}
\usepackage{caption}
\usepackage{subfigure}
\usepackage{makecell}
\usepackage{caption} 
\newcommand{\name}{\texttt{ENDEMIC}}
\newcommand{\dname}{\texttt{ECTF}}

\let\paragraph\oldparagraph
\let\subparagraph\oldsubparagraph

\usepackage{titlesec}
\usepackage{lipsum} 
\titlespacing*{\section}
{0pt}{2mm}{1mm}
\titlespacing*{\subsection}
{0pt}{1mm}{1mm}
\titlespacing*{\subsubsection}
{0pt}{1mm}{1mm}

\makeatletter
\newcommand{\printfnsymbol}[1]{%
  \textsuperscript{\@fnsymbol{#1}}%
}
\makeatother
\begin{document}



\title{Combining exogenous and endogenous signals with a semi-supervised co-attention network for early detection of COVID-19 fake tweets}

\author{\textbf{Rachit Bansal$^1$\thanks{Equal Contribution. The work was done when Rachit was an intern at IIIT-Delhi.}\quad William Scott Paka$^2$\printfnsymbol{1}\quad Nidhi$^3$\\ Shubhashis Sengupta$^3$\quad Tanmoy Chakraborty$^2$}}

\institute{$^1$ Delhi Technological University, { } $^2$ IIIT-Delhi, { } $^3$ Accenture Labs, India
\newline
\email{rachitbansal\_2k18ee152@dtu.ac.in \\ \{william18026, tanmoy\}@iiitd.ac.in \\ \{nidhi.sultan, shubhashis.sengupta\}@accenture.com}
}

\authorrunning{Bansal et al., 2021}
\titlerunning{ENDEMIC}

\maketitle

\begin{abstract}
Fake tweets are observed to be ever-increasing, demanding immediate countermeasures to combat their spread. During COVID-19, tweets with misinformation should be flagged and neutralised in their early stages to mitigate the damages. Most of the existing methods for early detection of fake news assume to have {\em enough propagation information} for {\em large labelled tweets} -- which may not be an ideal setting for cases like COVID-19 where both aspects are largely absent. In this work, we present \name, a novel early detection model which leverages exogenous and endogenous signals related to tweets, while learning on limited labelled data. We first develop a novel dataset, called \dname\ for early COVID-19 Twitter fake news, with additional behavioural test-sets to validate early detection.
We build a heterogeneous graph with follower-followee, user-tweet, and tweet-retweet connections and train a graph embedding model to aggregate propagation information. Graph embeddings and contextual features constitute endogenous, while time-relative web-scraped information constitutes exogenous signals.
\name\ is trained in a semi-supervised fashion, overcoming the challenge of limited labelled data. We propose a co-attention mechanism to fuse signal representations optimally. Experimental results on \dname, \textit{PolitiFact}, and \textit{GossipCop} show that \name\ is highly reliable in detecting early fake tweets, outperforming nine state-of-the-art methods significantly.


\end{abstract}

\section{Introduction}

Over the past couple of years, several social networking platforms have seen drastic increase in the number of users and their activities online \cite{bruna2013spectral}. 
Due to the lockdown situations and work from home conditions during COVID-19 pandemic, the screen time on social media platforms is at an all time high. Twitter is one such micro-blogging platform where users share opinions and even rely on  news updates. Twitter users exposed to unverified information and opinions of others often get influenced and become contributors to further spreading. The characteristics of fake news spreading faster farther and deeper than genuine news is well-studied \cite{lohr2018s}. Fake news on health during COVID-19 might endanger people's lives as a few of them call for action. Although our proposed method is highly generalised, we choose to specifically focus on the ongoing pandemic situation of COVID-19 as it is timely and needs scaleable solutions.

State-of-the-art fake news detection models for Twitter have proved useful when trained on a sufficient amounts of labelled data. Any emerging fake news in its initial stages could not be detected by such models due to the lack of corresponding labelled data. Moreover, these models tend to fail when the fake news is not represented largely in the training set. By the time the fake news is detected, it has spread and caused damages to a wide range of users. Detecting a fake news in an early stage of its spread gives us the advantage of flagging it early. 
A few state-of-the-art models for early fake news detection use propagation based methods \cite{Liu2018EarlyDO}, i.e., they are based on how a particular news event (fake/genuine) spreads (both wider and deeper); retweet chain in Twitter is one such example. 
These models work with the propagation chains and require sufficient historical data (retweet/reply cascade) of each tweet, which are very hard to collect within limited time.

In this work, we design \name, a novel  approach to detect fake news in its early stage. To this end, we developed an Early COVID-19 Twitter Fake news dataset, called \dname\ with additional test set (early-test) for early detection. We collected vast amount of COVID-19  tweets and label them using trusted information, transformer models and human annotation. We finally curate a standard training dataset, composed of numerous rumour clusters, while simulating a scenario of emerging fake news through early-test set. Next, we extract exogenous signals in form of most suitable stances from relevant external web domains, relative to the time the tweet is posted. Unlike existing studies which extract  propagation paths per tweet, we create a massive heterogeneous graph with follower-followee, tweet-retweet and tweet-user connections and obtain the node representations in an unsupervised manner.
The graph embeddings (both users and tweets) constitute the major component of the endogenous signals (within Twitter). 
The time-variant contextual tweet and user features are used to provide additional context, and a few of them are masked to validate the model for early detection. Lastly, to overcome the challenge of limited labelled dataset, we setup the whole model in a semi-supervised fashion, learning in an adversarial setting to utilise the vast unlabelled data.

In Summary, fake news on Twitter is an inevitable threat especially during  COVID-19 pandemic, where inaccurate or deviating medical information could be harmful. As a result, a timely model which can detect fake news in its early stages is important due to the current conditions of emerging fake news. We propose \name, a semi-supervised co-attention network  which utilises both exogenous and endogenous signals. Experimental results show that \name\ outperforms nine  state-of-the-art models in the task of early fake tweet detection. \name\ produces  93.7\% accuracy, on \dname\ for fake tweet detection and 91.8\% accuracy for early detection, outperforming baselines significantly. We also show the generalisation of \name\ by showing its efficacy on two publicly available fake news datasets, \textit{PolitiFact} and \textit{GossipCop}; \name\ achieves  91.2\% and 84.7\& accuracy on the two datasets, respectively.


In particular, our major contributions are as follows:
\begin{itemize}
    \item We introduce \dname, an  Early COVID-19 Twitter Fake news dataset with additional early-test set to evaluate the performance of fake tweet detection models for early detection.
    \item As propagation paths for tweets might not be available in all scenarios, we build connections of follower-followee, tweet-retweet and user-tweet network and extract representations upon learning the graph, constituting of our endogenous signals.
    \item We use exogenous signals which informs the model on the realities of information available on the web at tweet time, and helps in learning from weak external signals.
    \item Adding an effort towards early detection, time variant features are masked at test time for further evaluation on early detection.
    \item We further show the generalisation of \name\ by presenting its superior performance on two other general fake news datasets .
\end{itemize}
{\bf Reproducibility:} The code and the datasets are public at: \url{https://github.com/LCS2-IIITD/ENDEMIC}. 

\section{Related Work}
Our work focuses majorly on early detection of fake tweets. As our model involves techniques such as graphs and semi-supervised learning, we present related studies pertaining to our model.

\subsubsection*{Fake news detection.}
Proliferation of fake news over the Internet has given rise to many research, and hence there exist an abundance of literature. Early studies on fake news relied on linguistic features of texts to detect if the content is fake \cite{castillo2011information}. Wang et al. \cite{wang2020weak} used textual embeddings and weak labels with reinforcement learning to detect fake news. They leveraged an annotator component to obtain fresh and high-quality labelled samples to overcome the challenge of limited labelled data. Recent approaches have started exploring directed graph and Weisfeiler-Lehman Kernel based model for social media dataset that use similarity between different graph kernels \cite{rosenfeld2020kernel}. A recent survey \cite{zhou2020survey} shows that there are four approaches to detect fake news: (i) knowledge-based methods, where the content is verified with known facts, (ii) style-based methods, by analysing the writing style of the content, (iii) propagation methods, based on how a particular news event  spreads, and (iv)  source-based methods, by verification of credibility of sources. As studies show, each of these methods used individually is not enough to build an efficient classifier \cite{monti2019fake}.

\subsubsection*{Semi-supervised learning.}
Semi-supervised models have been used often in the past to leverage vast unlabelled datasets in various fields. 
 Helmstetter et al. \cite{helmstetter2018weakly}   explored weakly supervised learning for fake news detection which automatically collects large scale noisy dataset to aid the classification task . 
Yu et al. \cite{yu2017constrained} used constrained semi-supervised learning for social media spammer detection. Tensor embeddings are used to design a semi-supervised model for content based fake news detection \cite{guacho2018semi}. A few studies leveraged variational auto-encoders in the form of sequence-to-sequence modelling on text classification and sequential labelling \cite{gururangan2019variational}. Nigam et al. \cite{nigam2000text} classified the text using a combination of Naive Bayes and Expectation Maximisation algorithms and demonstrated substantial performance improvements. Miyato et al. \cite{miyato2016adversarial} utilised adversarial and virtual adversarial training to the text domain by applying perturbations to the word embeddings. Chen et al. \cite{chen2020mixtext} introduced MixText that combines labelled, unlabelled and augmented data for the task of text classification.

\subsubsection*{Early detection.} 
Early fake news detection methods detect fake news at an initial stage where the news has not yet been popularised. There exist very limited studies on early detection. Liu et al. \cite{Liu2018EarlyDO} built an early detection model using propagation paths of news spreading as a multivariate time series and then training a recurrent and convolution classifier on user features. Rosenfeld et al. used graph kernels on Twitter cascades to capture intricate details of the data-set for fake news detection without feeding user identity and time for an early detection model \cite{rosenfeld2020kernel}. Shu et al. used content engagement and cleaned labelled dataset  for early detection with deep neural network \cite{shu2018understanding}.

\section{Our Proposed Dataset: {\textbf{\dname}}}
\label{sec:dataset}

As there is no publicly available COVID-19 Twitter dataset particularly for {\em early detection}, we attempt to develop our own dataset, \dname, specifically crafting it for early detection. We expand on \texttt{CTF}, a general COVID-19 Twitter fake news dataset, proposed by Paka et al. \cite{paka2021cross}. \texttt{CTF} was formed using multiple sources- unlabelled Twitter datasets that are publicly released \cite{kaggleCarl, kaggleShane, kaggleSVEN}, hydration of tweets using predefined hashtags, and governmental health organisations and fact checking websites for verified news. They considered statements released by the latter to be true, and applied Sentence-BERT \cite{reimers-gurevych-2019-sentence} and RoBERTa \cite{liu2019roberta} to convert these tweets and verified news into contextual embeddings, pairwise cosine similarity is then computed to assign a label `fake' or `genuine. This way, \texttt{CTF} composes of $72,578$ labelled and $2,59,469$ unlabelled tweets, partially verified manually.\\
We took a sample of labelled and unlabelled tweets from \texttt{CTF}, forming our train set. A \textbf{`general-test'} set is created by randomly sampling from the remaining labelled and unlabelled tweets. The training dataset, which contains a wide variety of rumour clusters, is kept constant. We identified small rumour clusters in the labelled dataset and use those to form additional test set, called \textbf{`early-test'} set, to perform behavioural analysis of our algorithm for early detection. These small rumour clusters contain the fake news that are not popularised yet (having chance of getting popular), simulating early stages of fake news events. We extracted more tweets belonging to these rumour clusters, while keeping an upper limit on the time since they were posted. This ensures that the tweets belonging to this early-test set are in their early stages.

We refer to the complete dataset composed of the train and test sets as \textbf{\dname}, {\bf E}arly {\bf C}OVID-19 {\bf T}witter {\bf F}ake news dataset.

\section{Our Proposed Methodology: \name}
\begin{figure}[!t]
    \centering
    \includegraphics[width=0.9\textwidth]{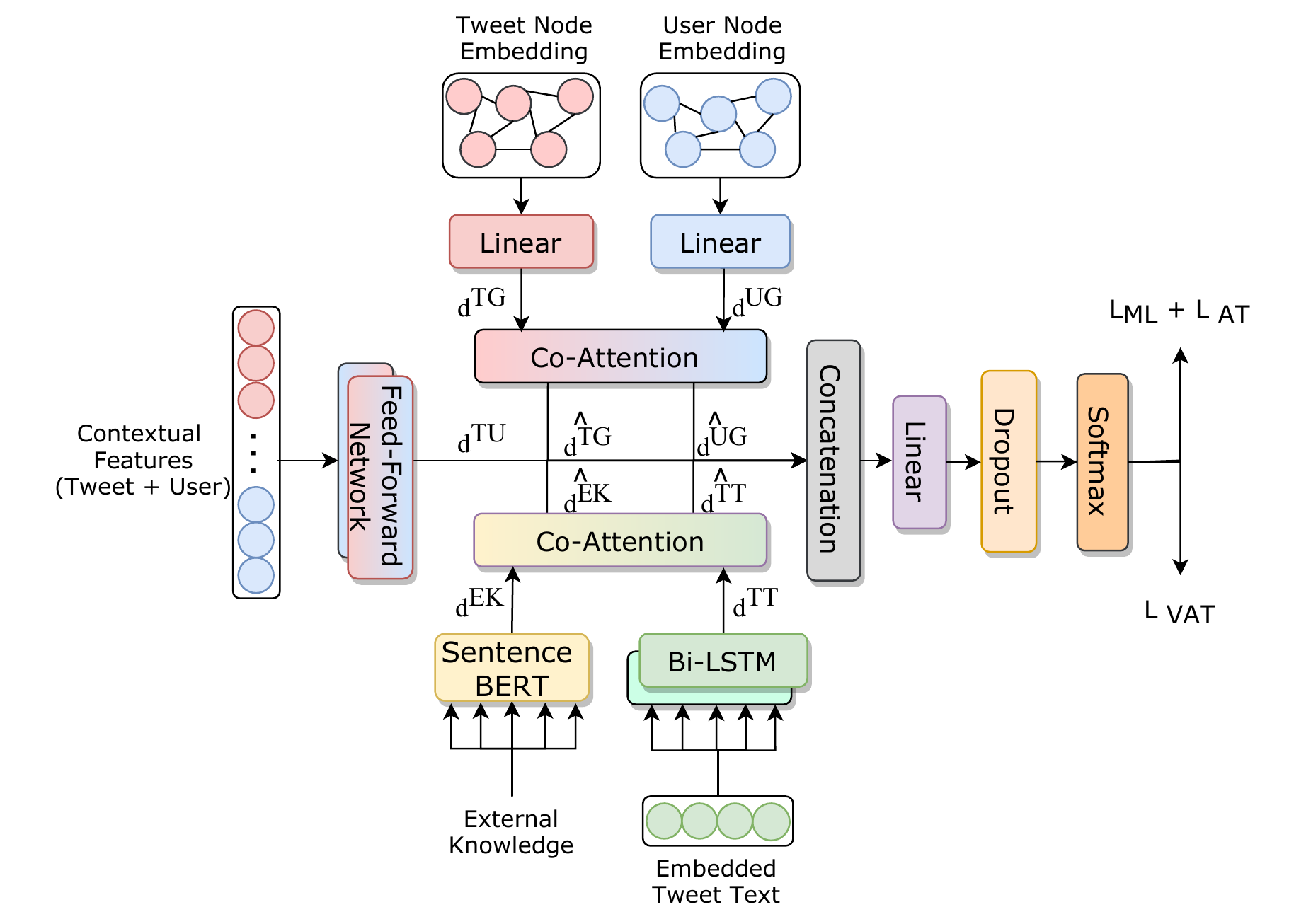}
    \setlength{\belowcaptionskip}{-7mm}
    \caption{A schematic architecture of \name. The encoded interpolations of external knowledge ($d^{EK}$), tweet text ($d^{TT}$), contextual features ($d^{TU}$), and tweet ($d^{TG}$) and user node embeddings ($d^{UG}$) are shown. $\widehat{\cdot}$\  represents an output from co-attention. 
    }
    \label{fig:arch}
\end{figure}

Here we present our model, called \textbf{\name} ({\bf E}xogenous and e{\bf ND}ogenous signals with s{\bf EMI}-supervised {\bf C}o-attention network).
Figure \ref{fig:arch} represents the model architecture. Here we show various input features, each of which passes through separate modules before being concatenated. Our approach towards early detection relies on simulating the early stages of the news event testing with time invariant features. 

\subsection{Exogenous signals}
Traditional supervised fake news detection methods are bounded by the knowledge present in the data they are trained on \cite{zhou_theory}. This makes them incapable of classifying tweets on information the model is not trained on. This is particularly challenging for domains where large amount of new theories and their consequent studies arrive rapidly within small duration. Very often, the general stance of such news also changes radically over time. Just as a human expert, classification models too need to be well-aware of current information in a domain in order to be reliable and efficient. To address this problem, we make use of exogenous signals through {\em external knowledge}.

We curate exogenous content using web-scraped articles from various sources. In order to simulate the early stages of detection, this knowledge base is collected differently for training and testing instances.
We build the external knowledge for training as a static knowledge base composed of various sources for each input. We hypothesise that collecting this external knowledge in accordance with the time the post was made simulates an early detection scenario, consequently making the model adapt to such scenarios better where the information about an emerging fake news will be limited. Using this, the model learns the weak signals that are present to perform a classification. In case of testing, the external knowledge is scraped for every test instance at the time of testing itself. The dynamic nature of building the external knowledge base during testing ensures that the model makes use of latest sources to make the prediction.

We perform a web scraping with tweets as queries using Google Web API. Since a large amount of information in form of web pages are available per query, and the content per page could also be excessively large, selecting the right content is vital. We do so by firstly tokenizing each web page $x^{EK,i}$ into sentences. The $j^{th}$ sentence of  $x^{EK,i}$, denoted by $x_j^{EK,i}$, is encoded using Sentence-BERT \cite{reimers-gurevych-2019-sentence} to obtain its contextual representation, $d_j^{EK,i} \in \mathbb{R}^K$, where $K$ is the dimension of the encoding. This representation is then compared with the representation of the tweet text, $x^{TT} \rightarrow d_x^{TT} \in \mathbb{R}^K$, encoded using the same Sentence-BERT model. This comparison is made using cosine similarity. If the cosine similarity between these two encodings, $cos(d_j^{EK,i} || d_x^{TT})$, is greater than a threshold $\epsilon$ (set as 0.8), then the sentence $x_j^{EK,i}$ is added to the set of input external knowledge for that particular tweet. The same representation, $d_j^{EK,i}$ is used as the corresponding input to the model for all sentences belonging to the tweet. This process is done for the entire set of input queries, until we obtain $50$ such sentences for each, with the amount of phrases per web-source being limited to $10$.
Thus, the net external knowledge input to \name\ during training is obtained by concatenating the encoding for each input in 2D fashion and is given by $d^{EK} \in \mathbb{R}^{n \times 50 \times K}$, where $n$ is the input size. For most of our experiments, we keep $K=512$.

\subsection{Endogenous signals}
\subsubsection{Input tweet embedding.}
The original tweet text $x_{i}^{TT}$ of sequence length $N$ (say) is first represented by a one-hot vector using a vocabulary of size $V$. A word embedding look-up table transforms these one-hot vectors into a dense tensor. This embedding vector is further encoded using a Bidirectional LSTM. A final state output $d \in \mathbb{R}^{K/2}$ is obtained at both the forward and backward layers, which are then concatenated to give a 2D vector corresponding to each text input. The final representation of the tweet text input to the model is, thus, $d^{TT} \in \mathbb{R}^{n \times N \times K}$.

\subsubsection{Graph embeddings.}
Extracting the propagation chain of retweets has been proven effective by the existing early detection models \cite{lu-li-2020-gcan, Liu2018EarlyDO}. However, these methods limit the training as for each training or test sample, the entire set of retweet chains needs to be extracted which is often computationally expensive. Here we build a heterogeneous graph $\mathcal{G}(\mathcal{V},\mathcal{E})$ as follows: the set of nodes $V$ can be users or tweets, and edges $E$ are formed in the following ways: two users are connected via follower-followee link, a user is connected to her posted tweet, two tweets are connected if one is a retweet of another. We keep $\mathcal{G}$ as undirected intentionally ignoring the direction of some edge types (follower-followee) in order to maintain uniformity. The formed heterogeneous graph contains around $51M$ nodes and $70M$ edges. Such huge connections, when added into one graph, form many disconnected clusters. 
In our graph, we observe one giant cluster with almost millions of nodes and edges, which stands dominating compared to other small (and disconnected) clusters.

We obtain the embedding of the graph using GraphSAGE \cite{niepert2016learning} in an unsupervised fashion
due to its capability to scale and learn large graphs easily. We label each node with its characteristics such as parent, tweet, retweet, user, fake tweet. The generalisability of GraphSAGE helps in extracting the embeddings of unseen users and tweets. We use a teleportation probability of $0.3$ for the graph to randomly jump across nodes, which helps with clusters having less number of connections or being disconnected from the rest. In this work, we show that using the embeddings of both users and tweets in combination with co-attention leads to better supervision and learning of the model.

We represent the tweet and user graph embeddings as $d^{TG}$ and $d^{UG}$ $\in \mathbb{R}^{n \times G}$ respectively, where $G$ represents the embedding dimension, and is kept as $G = 768$, for most of our experiments.

\subsubsection{Contextual features.}
Social media platforms like Twitter offer a variety of additional features  that can play a crucial role in identifying the general sentiment and stance on a post by the users. Moreover, some features of user could also be used as indicative measures of their social role and responsibility, in general. Therefore, we use a variety of such tweet and user features to provide additional context regarding the input tweet and the corresponding users who interact with it. Some of the tweet features used are \textit{number of favourites, number of retweets, PageRank reliability score of domains mentioned in the tweet}, \textit{text harmonic sentiment}\footnote{obtained from the \href{https://textblob.readthedocs.io/en/dev/}{textblob} tool in Python.}, etc. Some of the user features include \textit{follower and followee counts}, {\em verified status}, and \textit{number of tweets made by the user}.

Note that majority of these features continually change over time and can be regarded as time-variant and accordingly, the inferences drawn also would change over time. Therefore, for early detection, it is vital not to rely too heavily on such features. For instance, the number of likes and retweets for a tweet changes over time. Similarly, such additional features for a new user cannot be expected to give a proper indicative measure of a user's tendency of believing, spreading, or curbing misinformation. And masking the time-variance during evaluation is better explained in Section \ref{sec:early_1}.

Throughout this study, we represent these contextual tweet and user features as $x^{TF} \in \mathbb{R}^{n \times N_{TF}}$ and $x^{TF} \in \mathbb{R}^{n \times N_{UF}} $, respectively, where $N_{TF}$ and $N_{UF}$ indicate the number of such features. As shown in Fig. \ref{fig:arch}, these input features are concatenated and passed across a common feed-forward network (FFN), which interpolates $x_i^{TF} \oplus x_i^{UF} \in \mathbb{R}^{1 \times (N_{TF}+N_{UF})}$ to $d_i^{TU} \in \mathbb{R}^{C}$, where $C$ is the output dimension of FFN.

\subsection{Connecting components and training} 

\subsubsection{Co-attention.}

\begin{wrapfigure}{r}{0.45\textwidth}
    \centering
    \includegraphics[width=0.4\textwidth]{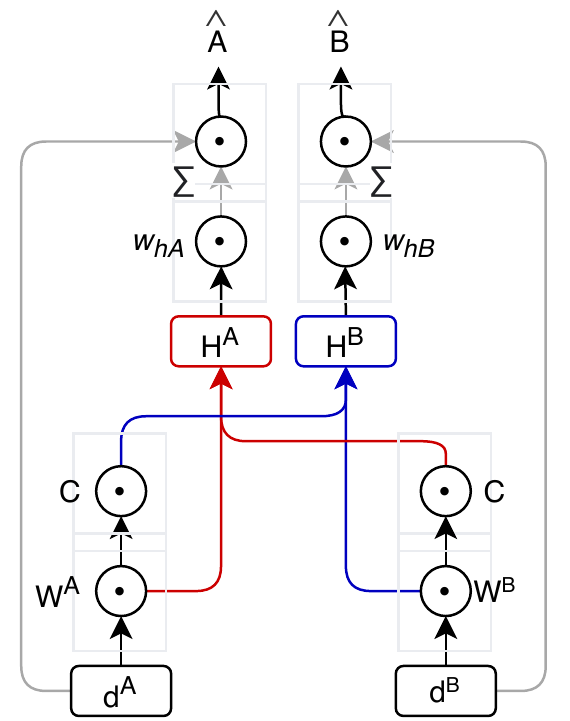}
    \caption{The co-attention mechanism.}
    \label{fig:co_attn}
\end{wrapfigure}

In order to  jointly attend and reason about various interpolated inputs, we use the parallel co-attention mechanism \cite{co_attn_yang} (Figure \ref{fig:co_attn}). As shown in Figure \ref{fig:arch}, this is done at two places of \name, namely, to co-attend between external knowledge $d_{i}^{EK} \in \mathbb{R}^{50 \times K}$ and tweet text   $d^{TT} \in \mathbb{R}^{N \times K}$ in 2D, and tweet $d^{TG} \in \mathbb{R}^{1 \times G}$ and user $d^{UG} \in \mathbb{R}^{1 \times G}$ graph embeddings in 1D. The same process is followed for two; therefore, to unify the notation, we use $d^{A} \in \mathbb{R}^{X \times Z}$ and $d^{B} \in \mathbb{R}^{Y \times Z}$ to explain the mechanism used.

Firstly, an affinity matrix $C \in \mathbb{R}^{Y \times X}$ is obtained as, $C = tanh(d^{A}W_{b}{d^{B}}^\top)$. Here $W_{b} \in \mathbb{R}^{Z,Z}$ represents the learnable weight matrix. Further the corresponding attention maps between $A$ and $B$ are calculated as follows:
\begin{equation*}
    \begin{aligned}
        H^A = tanh(W_{A}{d^{A}}^\top + (W_{B}{d^{B}}^\top)C), \hspace{0.3cm} H^B = tanh(W_{B}{d^{B}}^\top + (W_{A}{d^{A}}^\top)C^\top)
    \end{aligned}
\end{equation*}
where, $W_A$, $W_B$ $\in \mathbb{R}^{k \times Z}$ again represent the learnable weight matrices. Further, to compute the attention probabilities of each element in $A$ with each element of $B$, we use,
\begin{equation*}
    \begin{aligned}
        a^{A} = Softmax({w_{hA}}^{\top}H^A), \hspace{0.3cm} a^{B} = Softmax({w_{hB}}^{\top}H^B)
    \end{aligned}
\end{equation*}
where, $w_{hA}$, $w_{hB}$ $\in \mathbb{R}^{k}$ represent the weight parameters, while $a^{A} \in \mathbb{R}^X$ and $a^B \in \mathbb{R}^Y$ represent the resultant attention probabilities. Finally, the net attention vectors between $A$ and $B$ are computed as a weighted sum of the corresponding features, i.e., 

\begin{equation}
    \begin{aligned}
        \widehat{A} = \sum_{i=1}^{X} a_{i}^{A}d_{i}^{A}, \hspace{0.3cm} \widehat{B} = \sum_{i=1}^{Y} a_{i}^{B}d_{i}^{B}
    \end{aligned}
\end{equation}

$\widehat{A}$ and $\widehat{B}$ $\in \mathbb{R}^{1 \times Z}$ are the learned feature vectors obtained through co-attention. This, for instance, represents how each representation in the tweet graph embeddings attends to the user graph embeddings, when $A$ represents tweet graph embeddings ($TG$), and $B$ represents user graph embeddings ($UG$).

The interpolations of the various inputs, obtained through the separate smaller pipelines are combined using a unified head pipeline in the model architecture. Considering a single input, firstly, the representations from the two co-attention layers, $\widehat{d^{EK}}$ and $\widehat{d^{TT}}$ $\in \mathbb{R}^{1 \times K}$, and $\widehat{d^{TG}}$ \& $\widehat{d^{UG}}$ $\in \mathbb{R}^{1 \times G}$, are concatenated along with $d^{TU} \in \mathbb{R}^{1 \times C}$.

\begin{equation*}
    d = \widehat{d^{EK}} \oplus \widehat{d^{TT}} \oplus \widehat{d^{TG}} \oplus \widehat{d^{UG}} \oplus d^{TU}
\end{equation*}

This net representation, $d \in \mathbb{R}^{2K + 2G + C}$, then passes onto a dropout ($p_{drop} = 0.2$) regularised feed-forward layer with an output size of $2$, and finally a Softmax function, to give the probability per output class.

\subsubsection{Training.}
In order to overcome the limitations produced by scarce labelled data, we use the Virtual Adversarial Loss (VAT) across both labelled and unlabelled data. VAT introduces a small perturbation in the input embedding and computes the loss using weak labels. Maximum Likelihood (ML) loss and Adversarial Training (AT) losses are further used to train \name\ on the labelled data \cite{miyato2016adversarial}. These additional losses allow \name\ to be more robust, and the abundant unlabelled data being used this way allow it to form better understanding and representation of the domain-specific texts. 


\section{Experimental Setup and Results}
Here we present our comparative analysis by empirically comparing \name\ with  state-of-the-art models, including techniques for general fake news detection, early detection, and text classification.

\subsection{Baseline methods} We consider nine baselines as follows. Liu and Wu \cite{Liu2018EarlyDO} introduced a GRU and CNN-based model which relies on the propagation path (PP) of a news source to classify its veracity, and is termed as \textbf{PPC} (Propagation Path Classification). \textbf{FNED} \cite{fned} makes use of a PU-learning framework to perform weak classification, and is claimed to be suitable in data scarce settings with large amount of unlabelled data, as is the scenario we deal with in this study. Further, Shu et al. \cite{shu2020leveraging} proposed to learn from Multiple-sources of weak Social Supervision (\textbf{MWSS}), on top of a classifier such as CNN and RoBERTa \cite{liu2019roberta}. They too relied on weak labelling, but constructed them through social engagements. Both \textbf{dEFEND} and \textbf{CSI} deployed an RNN-based framework for classifying news; the former makes use of user comments with co-attention, while the latter relies on a three-step process of \textit{capturing, scoring} and \textit{integrating} the general user response. Furthermore, \textbf{GCAN} \cite{lu-li-2020-gcan} presents a dual co-attention mechanism over source and retweet tweet representations, relying on interaction and propagation for detecting fake news. Finally, we also make some interesting observations by employing text classification models, namely \textbf{MixText} \cite{chen2020mixtext} and \textbf{HAN} \cite{yang2016hierarchical}, as additional baselines. 

\subsection{Evaluating on general-test set}
\begin{table}[!t]
\centering
    \renewcommand{\arraystretch}{1.05}
    \scalebox{0.80}{
    \begin{tabular}{ l||c|c|c|c||c|c|c|c}
        \hline
        \multirow{2}{*}{{\bf Model}} &\multicolumn{4}{p{2cm}||}{{\bf Feature used}} &\multicolumn{4}{c}{{\bf Performance}} \\
        \cline{2-9}
        &TG &UG &EK &UL &Accuracy &Precision & Recall &F1 Score \\\hline
        \hline
        HAN & & & & &0.865 &0.685 &0.865 &0.762 \\
        \hline
        MixText & & & &\checkmark &0.875 &0.835 &0.843 &0.847 \\
        \hline
        GCAN & &\checkmark & & &0.852 &0.817 &0.820 &0.813 \\
        \hline
        CSI &\checkmark & &\checkmark & &0.873 &0.805 &0.915 &0.854 \\
        \hline
        dEFEND &\checkmark & &\checkmark & &0.890 &0.830 &0.891 &0.862 \\
        \hline
        CNN-MWSS & & & &\checkmark &0.805 &0.785 &0.810 &0.790 \\
        \hline
        RoBERTa-MWSS & & & &\checkmark &0.825 &0.815 &0.852 &0.840 \\
        \hline
        FNED & & & &\checkmark &0.911 &0.907 &0.920 &0.915 \\
        \hline
        PPC &\checkmark & & & &0.861 &0.860 &0.832 &0.845 \\
        \hline
        \textbf{\name} &\checkmark &\checkmark &\checkmark &\checkmark &\textbf{0.937} &\textbf{0.922} &\textbf{0.943} &\textbf{0.932}\\
        \hline
    \end{tabular}}
\setlength{\belowcaptionskip}{-6mm}
\caption{Features used and  performance comparison on {\bf general-test} dataset of \dname\ (TG: Tweet Graph, UG: User Graph, EK: External Knowledge, UL: Unlabelled Data).}
\label{tab:ress1}
\end{table}

Table \ref{tab:ress1} shows the performance comparison of \name\ compared to the baselines on the {\bf general-test} set of \dname. We observe that all the features (graph based, external knowledge and unlabelled data) play a major role in determining the corresponding performance of detecting fake tweets. While general fake news detection models like dEFEND and CSI strive to integrate more such features, early detection models like FNED and PPC tend to rely more on the time-invariant features like text alone. The effect produced by the absence of one over the other is apparent from Table \ref{tab:ress1}. In general, \name\ shows a benchmark performance of $0.937$ accuracy, outperforming all the baselines across all the evaluation measures significantly.

\subsection{Evaluating on early detection}
\label{sec:early_1}
\begin{table}[t]
\scalebox{0.845}{
\parbox{.5\linewidth}{
\centering
    \renewcommand{\arraystretch}{1.5}
    \begin{tabular}[H]{|l|c|c|c|c|c|}
     \hline
     \multirow{2}{*}{{\bf Model}} &\multicolumn{4}{c|}{{\bf Performance}} & \\
     \cline{2-6}
     &Acc. &Prec. &Rec. &F1 &$\Delta Acc$\\
     \hline
     HAN &0.850 &0.692 &0.841 &0.751 &1.5\% \\
     \hline
     MixText &0.865 &0.832 &0.851 &0.844 &1.0\% \\
     \hline
     GCAN &0.809 &0.800 &0.785 &0.795 &4.3\% \\
     \hline
     CSI &0.808 &0.821 &0.795 &0.815 &6.5\% \\
     \hline
     dEFEND &0.800 &0.790 &0.815 &0.790 &9.0\% \\
     \hline
     MWSS &0.798 &0.800 &0.812 &0.805 &2.7\% \\
     \hline
     FNED &0.901 &0.887 &0.913 &0.895 &1.0\% \\
     \hline
     PPC &0.831 &0.805 &0.797 &0.810 &3\% \\
     \hline
     \textbf{\name} &\textbf{0.918} &\textbf{0.910} &\textbf{0.923} &\textbf{0.920} &1.9\% \\
     \hline
    \end{tabular}
\setlength{\belowcaptionskip}{-8mm}
\caption{Comparing model performance on {\bf early-test} set of \dname.}
\label{tab:early_detect_1}
}}
\scalebox{0.8}{
\parbox{.65\linewidth}{
\centering
    \renewcommand{\arraystretch}{1.25}
    \begin{tabular}[H]{|l|c|c|c|c|c|c|}
     \hline
     \multirow{2}{*}{{\bf Model}} &{\bf Time-variant} &\multicolumn{4}{c|}{\bf Performance} & \\
     \cline{3-6}
     &{\bf Features} &Acc. &Prec. &Rec. &F1 &$\Delta Acc$\\
     \hline
     HAN &-- &0.858 &0.705 &0.850 &0.755 &1.2\% \\
     \hline
     MixText &-- &0.860 &0.815 &0.833 &0.820 &1.0\& \\
     \hline
     GCAN &\makecell{Tweet\\Propagation} &0.792 &0.812 &0.775 &0.805 &6.0\% \\
     \hline
     CSI &\makecell{User\\Response}&0.783 &0.701 &0.745 &0.725 &9.0\% \\
     \hline
     dEFEND &\makecell{User\\Comments}&0.768 &0.750 &0.804 &0.774 &12.2\% \\
     \hline
     MWSS &-- &0.810 &0.800 &0.835 &0.844 &1.5\% \\
     \hline
     FNED &-- &0.892 &0.890 &0.925 &0.905 &1.9\% \\
     \hline
     PPC &\makecell{Propagation\\Paths} &0.855 &0.810 &0.790 &0.805 &0.6\% \\
     \hline
     \textbf{\name} &\makecell{Contextual\\Features} &\textbf{0.928} &\textbf{0.920} &\textbf{0.930} &\textbf{0.927} &0.9\% \\
     \hline
    \end{tabular}
\setlength{\belowcaptionskip}{-8mm}
\caption{Performance comparison for {\bf mask-detect} on \dname. The masked features and corresponding metric scores are shown.}
\label{tab:early_detect_2}
}}
\end{table}

\subsubsection{Early-test.} Table \ref{tab:early_detect_1} shows the comparative results on the specially curated evaluation set ({\bf early-test}) for early detection of fake news.
Even though the change in accuracy ($\Delta Acc$) of \name\ as compared to its performance on general fake news (as shown in Table \ref{tab:ress1}) is not least among the rest of the models, it can be seen that it still comfortably outperforms all other baselines. Interestingly, the general purpose text classifiers, which are not particularly designed for fake news detection, show a relatively lesser $\Delta Acc$, while dEFEND \cite{cui2019defend} suffers from the largest difference of $9\%$. We attribute this to the heavy reliance of these models on time-variant features which provide critically less context in early stages. At the same time, as shown in Table \ref{tab:ress1}, only text-based and closely related non-time-variant features are not enough to reliably detect fake news. Thus, the set of input features used by \name\ optimises the trade-off between reliability and time-variance. 

\subsubsection{Masking time variance.}
\label{sec:masking}


To further verify our model's prowess in detecting early fake news, we introduce a unique masking approach to be applied upon the early-test evaluation set. For this technique, the tweets belonging to the smaller rumour clusters (as defined in Section \ref{sec:dataset}) are further simulated as actual early tweets for a model, by masking all time-variant features used by the model. We call this approach \textbf{`mask-detect'}.

Most of the existing techniques for fake news detection make use of some particular input features which are time-variant. In case of \name, these are the additional contextual tweet features and user features, which rapidly change over time. When such features are masked, there is effectively no way to distinguish an old tweet from a new one. Therefore, we perform mask-detect by replacing the numerical values of the relevant time-variant features with a common masking token. These features are different for each model. Therefore, we first identify such features and then perform the masking.
Table \ref{tab:early_detect_2} shows the features masked for the various fake news detection models and their corresponding performance.

\subsection{Evaluating on general-domain fake news}

\begin{table}[t]
\centering
    \renewcommand{\arraystretch}{1.25}
    \scalebox{0.8}{
    \begin{tabular}{ l||c|c|c|c||c|c|c|c}
        \hline
        \multirow{2}{*}{{\bf Model}} &\multicolumn{4}{c||}{{\bf PolitiFact}} &\multicolumn{4}{c}{{\bf GossipCop}} \\
        \cline{2-9}
        &Accuracy &Precision & Recall &F1 Score &Accuracy &Precision & Recall &F1 Score \\\hline
        \hline
        HAN &0.863 &0.730 &0.855 &0.790 &0.800 &0.685 &0.775 &0.770 \\
        \hline
        MixText &0.870 &0.840 &0.860 &0.853 &0.785 &0.805 &0.795 &0.813 \\
        \hline
        GCAN &0.867 &0.855 &0.880 &0.875 &0.790 &0.805 &0.795 &0.803 \\
        \hline
        CSI &0.827 &0.847 &0.897 &0.871 &0.772 &0.732 &0.638 &0.682 \\
        \hline
        dEFEND &0.904 &0.902 &\textbf{0.956} &\textbf{0.928} &0.808 &0.729 &0.782 &0.755 \\
        \hline
        CNN-MWSS &0.823 &0.811 &0.830 &0.820 &0.770 &0.793 &0.815 &0.795 \\
        \hline
        RoBERTa-MWSS &0.825 &0.833 &0.795 &0.805 &0.803 &0.815 &0.810 &0.807 \\
        \hline
        FNED &0.907 &\textbf{0.910} &0.922 &0.916 &0.832 &0.830 &0.825 &0.830 \\
        \hline
        PPC &0.885 &0.887 &0.870 &0.880 &0.791 &0.805 &0.832 &0.811 \\
        \hline
        \textbf{\name} &\textbf{0.912} &\textbf{0.910} &0.904 &0.915 &\textbf{0.847} &\textbf{0.835} &\textbf{0.822} &\textbf{0.840}\\
        \hline
    \end{tabular}}
\setlength{\belowcaptionskip}{-6mm}
\caption{Performance comparison on two general-domain datasets to check the generalisability of the models.}
\label{tab:ress2}
\end{table}

Although we tested our model on COVID-19, \name\ is highly generalised to other domains. And to prove the generalisability, we also evaluate \name\ on the FakeNewsNet \cite{shu2018fakenewsnet} datasets -- \textit{PolitiFact} and \textit{GossipCop}. Both of these datasets are obtained from the respective sources and contain labelled textual news content along with the social context. Evaluating on these datasets allows us to validate \name's generalisation across domains outside of COVID-19 and beyond \dname.

Table \ref{tab:ress2} shows that \name\ outperforms all baselines with more than 0.5\% accuracy on \textit{PolitiFact} and 1.5\% on \textit{GossipCop}. The performance of \name\ is highly comparable compared to other baselines  w.r.t. other evaluation metrics. 
These results clearly establish that \name\ generalises well on any fake news domain and dataset. 

\section{Conclusion}
In this work, we introduced the task of early detection of COVID-19 fake tweets. We developed a COVID-19 dataset with additional test set to evaluate models for early detection. We took measures to simulate the early stages of the fake news events by extracting the external knowledge relative to the time the tweet is posted. Our proposed model, \name\ overcomes the challenges of limited dataset  in a semi-supervised fashion. We adhere to co-attention as a better information fusion tested on time invariant features. \name\ outperformed nine baselines in both the tasks of general and early-stage fake news detection.
Experimental results compared to nine state-of-the-art models show that \name\ is highly capable for early detection task. We also showed the generalisability  of \name\ on two other publicly available fake news dataset which are not specific to COVID-19.

\section*{Acknowledgements}
The work was partially supported by Accenture Labs, SPARC (MHRD) and CAI, IIIT-Delhi. T. Chakraborty would like to thank the support of the Ramanujan Fellowship.

{\small
\bibliographystyle{endemic}
\bibliography{endemic}}

\end{document}